\documentclass[11pt]{article}
\usepackage[preprint]{acl}
\usepackage{times}
\usepackage{latexsym}
\usepackage[T1]{fontenc}
\usepackage[utf8]{inputenc}
\usepackage{microtype}
\usepackage{inconsolata}
\usepackage{graphicx}
\usepackage{amsmath,amssymb}
\usepackage{booktabs}
\usepackage{placeins}

\newcommand{\AgreeHalfLensO}{30}
\newcommand{\AgreeHalfLensS}{35}
\newcommand{\AgreeNineLRFO}{17}
\newcommand{\AgreeNineLRFS}{18}

\newcommand{\AngJLMidO}{77}

\newcommand{\AngLLMidO}{75}

\newcommand{\AngLensLateMinO}{41}

\newcommand{\AttrAttGradO}{0.84}
\newcommand{\AttrAttGradS}{0.75}
\newcommand{\AttrAttJLO}{0.20}

\newcommand{\AttrAttLRFO}{0.85}
\newcommand{\AttrAttLRFS}{0.79}

\newcommand{\AttrGradO}{0.68}
\newcommand{\AttrGradS}{0.62}

\newcommand{\AttrLRFO}{0.90}
\newcommand{\AttrLRFS}{0.88}
\newcommand{\AttrMlpGradO}{0.71}
\newcommand{\AttrMlpGradS}{0.63}

\newcommand{\AttrMlpLRFO}{0.90}
\newcommand{\AttrMlpLRFS}{0.88}

\newcommand{\AttrSeqsS}{64}

\newcommand{\BasisInputO}{0.00}
\newcommand{\BasisInputS}{0.42}

\newcommand{\BasisLabelsS}{64}

\newcommand{\BasisOrdOneO}{0.97}
\newcommand{\BasisOrdOneS}{0.84}

\newcommand{\BasisPulledO}{1.00}
\newcommand{\BasisPulledS}{0.98}

\newcommand{\BasisReadoutO}{0.79}

\newcommand{\BasisReadoutOrdOneO}{0.73}
\newcommand{\BasisReadoutOrdOneS}{0.54}
\newcommand{\BasisReadoutS}{0.56}

\newcommand{\BasisStatesS}{64}
\newcommand{\BasisStep}{0.1}

\newcommand{\BasisUnchangedO}{0.80}
\newcommand{\BasisUnchangedS}{0.45}

\newcommand{\CIKLMax}{12}

\newcommand{\CIShareMax}{0.04}

\newcommand{\CISteerSide}{0.03}

\newcommand{\CalHiO}{0.151}
\newcommand{\CalHiS}{0.163}
\newcommand{\CalLoO}{0.146}
\newcommand{\CalLoS}{0.160}
\newcommand{\CalibSeqs}{256}
\newcommand{\CkptsO}{9}

\newcommand{\CmpIPFrameKLO}{0.15}
\newcommand{\CmpIPFrameKLS}{0.16}

\newcommand{\CmpPairModelO}{0.02}
\newcommand{\CmpPairModelS}{0.10}

\newcommand{\CmpRowsS}{512}

\newcommand{\CosLRFMidO}{0.97}
\newcommand{\CosLRFMidS}{0.97}

\newcommand{\CostRatioO}{44}
\newcommand{\CostRatioS}{18}

\newcommand{\DirTokensO}{64}

\newcommand{\ECELRFO}{0.018}
\newcommand{\ECELRFS}{0.027}

\newcommand{\ECEModelO}{0.023}
\newcommand{\ECEModelS}{0.018}

\newcommand{\ExContext}{A review of Boulter 's performance in The}

\newcommand{\ExOneJL}{20}
\newcommand{\ExOneLL}{24}
\newcommand{\ExOneLRF}{16}

\newcommand{\ExProb}{0.29}
\newcommand{\ExProbRunnerUp}{0.22}
\newcommand{\ExRunnerUp}{``\tok{Times}''}
\newcommand{\ExRunnerUpLayer}{12}

\newcommand{\ExToken}{``\tok{Guardian}''}

\newcommand{\GateTrueO}{0.0002}
\newcommand{\GateTrueS}{0.0009}
\newcommand{\GpuHours}{100}

\newcommand{\GpuHoursTotal}{500}
\newcommand{\JLPromptsO}{100}

\newcommand{\KLLLRandO}{0.52}

\newcommand{\KLLLTrainFirstO}{2.32}
\newcommand{\KLLLTrainFirstS}{4.82}
\newcommand{\KLLLTrainLastO}{5.66}
\newcommand{\KLLLTrainLastS}{29.06}

\newcommand{\KLLRFO}{0.15}
\newcommand{\KLLRFRandO}{0.11}
\newcommand{\KLLRFS}{0.16}
\newcommand{\KLLRFTrainFirstO}{0.02}
\newcommand{\KLLRFTrainFirstS}{0.15}
\newcommand{\KLLRFTrainLastO}{0.15}
\newcommand{\KLLRFTrainLastS}{0.16}

\newcommand{\KLLongLRFO}{0.18}
\newcommand{\KLLongLRFS}{0.25}
\newcommand{\KLLongTwoO}{0.08}
\newcommand{\KLLongTwoS}{0.15}

\newcommand{\KLShortLRFO}{0.15}
\newcommand{\KLShortLRFS}{0.16}
\newcommand{\KLShortTwoO}{0.05}
\newcommand{\KLShortTwoS}{0.06}

\newcommand{\KLTwoO}{0.05}
\newcommand{\KLTwoS}{0.06}

\newcommand{\KLZeroO}{0.93}
\newcommand{\KLZeroS}{0.87}

\newcommand{\MidLayerO}{16}

\newcommand{\NumLayersO}{32}
\newcommand{\NumLayersS}{36}

\newcommand{\PlateauFirstS}{8.1}
\newcommand{\PlateauFromS}{8}
\newcommand{\PlateauLastAbsentS}{2.5}
\newcommand{\PlateauToS}{17}
\newcommand{\PlateauValS}{0.29}

\newcommand{\ReadSeqsS}{64}
\newcommand{\RelaxAboveToO}{14}
\newcommand{\RelaxAboveToS}{17}
\newcommand{\RelaxRateO}{10.5}
\newcommand{\RelaxRateS}{9.2}
\newcommand{\RelaxRsqO}{0.99}
\newcommand{\RelaxRsqS}{0.98}
\newcommand{\SeqLen}{128}

\newcommand{\SettleExitLLO}{0.04}

\newcommand{\SettleExitLRFO}{0.42}
\newcommand{\SettleExitLRFS}{0.45}

\newcommand{\SettleJLO}{0.09}
\newcommand{\SettleJLS}{0.09}
\newcommand{\SettleLLO}{0.11}
\newcommand{\SettleLLS}{0.03}
\newcommand{\SettleLRFO}{0.81}
\newcommand{\SettleLRFS}{0.82}

\newcommand{\SettleRowsS}{8128}

\newcommand{\SettleTwoO}{0.90}
\newcommand{\SettleTwoS}{0.89}

\newcommand{\SettleZeroO}{0.58}
\newcommand{\SettleZeroS}{0.58}
\newcommand{\SqrtKappaMidO}{0.35}
\newcommand{\SqrtKappaMidS}{0.30}

\newcommand{\SteerBasisLRFSelO}{0.82}
\newcommand{\SteerBasisLRFSelS}{0.71}

\newcommand{\SteerBasisSelO}{0.85}
\newcommand{\SteerBasisSelS}{0.73}
\newcommand{\SteerBasisSideO}{0.06}
\newcommand{\SteerBasisSideS}{0.11}

\newcommand{\SteerLLReachO}{0.81}

\newcommand{\SteerLRFReachO}{0.94}
\newcommand{\SteerLRFReachS}{0.92}

\newcommand{\SteerNativeReachO}{0.97}
\newcommand{\SteerNativeReachS}{0.95}

\newcommand{\SteerNativeSideO}{0.19}
\newcommand{\SteerNativeSideS}{0.22}

\newcommand{\SubCloudHi}{0.28}

\newcommand{\SubJLS}{0.28}

\newcommand{\SubLRFFirstO}{0.66}
\newcommand{\SubLRFFirstS}{0.64}

\newcommand{\SubLRFO}{0.95}
\newcommand{\SubLRFS}{0.94}

\newcommand{\SubPairsS}{7961}

\newcommand{\TopLLTrainFirstO}{0.27}
\newcommand{\TopLLTrainFirstS}{0.21}
\newcommand{\TopLLTrainLastO}{0.15}
\newcommand{\TopLLTrainLastS}{0.04}
\newcommand{\TopLRFFirstO}{0.39}
\newcommand{\TopLRFFirstS}{0.42}
\newcommand{\TopLRFO}{0.86}
\newcommand{\TopLRFS}{0.86}

\newcommand{\TopLRFTrainLoO}{0.86}
\newcommand{\TopLRFTrainLoS}{0.85}

\newcommand{\tok}[1]{{\small\ttfamily #1}}  
\newcommand{\lrfn}[1]{{\small\mbox{\textsc{LexFrame}[#1]}}}
\newcommand{\lrf}{\lrfn{1}}

\title{Lexical Reference Frames:\\ Interpreting LM Residual Streams in Native Semantics}

\author{Jian Gu \\
  Monash University \\
  \texttt{jian.gu@monash.edu}
\\\And
  Hongyu Zhang \\
  Chongqing University \\
  \texttt{hyzhang@cqu.edu.cn}
\\\And
  Chunyang Chen \\
  Technical University of Munich \\
  \texttt{chun-yang.chen@tum.de}
\\\AND
  Aldeida Aleti \\
  Monash University \\
  \texttt{aldeida.aleti@monash.edu}
}

\begin{document}
\maketitle

\begin{abstract}
The vocabulary of a language model is its native semantics, and the embedding and unembedding fix the meaning of a state at the two ends of the residual stream.
Between them, existing methods reuse a fixed lens or treat states as a point cloud, and neither reads a state as the model does.
We argue that every hidden state has a native reference frame, induced by the layers that remain and adaptive to the context.
Lexical reference frames approximate it in a model-native and context-aware way, reading a state through the remaining layers with a given number of rounds of attention between positions and a typical attention output for the rest, fitting nothing to the model's output.
No lens with fixed directions per layer can reach a mean cosine with the native frame's lexical readouts above the square root of their coherence, \SqrtKappaMidS{} and \SqrtKappaMidO{} at mid-depth of SmolLM3-3B and OLMo3-7B. \lrf{}, of order 1, reaches \CosLRFMidS{} and \CosLRFMidO{}, and its readings are on average within \KLLRFS{} and \KLLRFO{} nats of the model's output.
Its coefficient differences attribute predictions more faithfully than attribution patching, and its lexical basis steers predictions while sparing other tokens. On average, its readings settle on the model's top token over the final \SettleLRFS{} and \SettleLRFO{} of the layers, those of lenses over at most \SettleLLO{}.
These frames thus make every hidden state readable, attributable and steerable.\footnote{\url{https://github.com/jianguda/lexframe}}
\end{abstract}

\section{Introduction}

Between its input and output tokens, a language model computes nothing but states of its residual stream, which every layer reads and writes \citep{elhage2021mathematical}.
Its vocabulary is in this sense the native semantics of the model, the smallest units of meaning that it receives and produces.
The model defines the lexical meaning of a state at the two ends of the stream. The embedding assigns each token a state, and the unembedding maps a final state to logits over tokens.
Intermediate states lack such a definition, although interpretability research and its applications read them constantly. Early exit and self-speculative decoding stop the computation at a layer whose reading already names the output \citep{schuster2022calm,elhoushi2024layerskip}, hallucination detectors judge an answer by its hidden states \citep{orgad2025llms,zhang2025icr} and knowledge editors locate and rewrite facts at intermediate layers \citep{meng2023memit,fang2025alphaedit}.
Each depends on how faithfully a lens or a similar tool reads these states. Model compression also relies on what these states mean, as it distils a smaller or pruned student by matching the intermediate states of its teacher \citep{romero2015fitnets,dasgupta2025hidden,muralidharan2024compact}.
This meaning is set by the remaining layers, which read each state in the context of its prefix.

\begin{figure*}[htb]
\centering
\includegraphics[width=\textwidth]{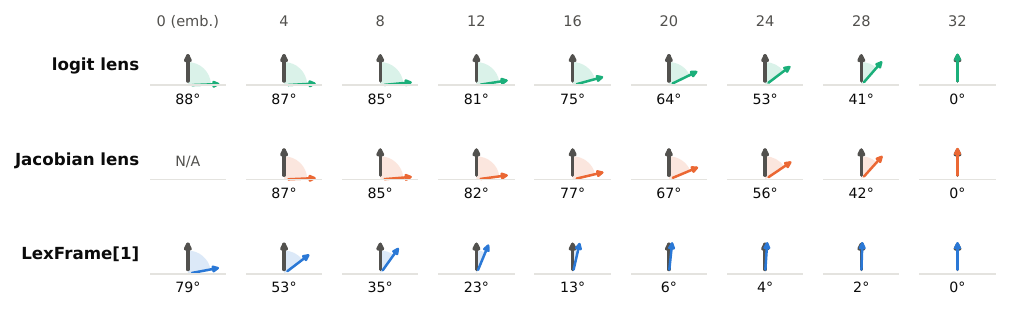}
\caption{Angle between the lexical readout of \lrfn{$\infty$} for a label (grey) and the direction each method offers, after every fourth layer of OLMo3-7B, as means over states and labels, zero at the output by definition.}
\label{fig:rulers}
\end{figure*}

Two families of methods address these states and fail in opposite ways.
The first reads a state with a lens. The logit lens applies the output frame to every layer unchanged \citep{nostalgebraist2020logitlens}, the tuned lens learns one affine map per layer fitted to the model's output \citep{belrose2023tuned} and the Jacobian lens averages the Jacobian of the remaining layers over positions and prompts \citep{gurnee2026workspace}.
Each fixes one linear map per layer, so every state of a layer is read identically, regardless of its prefix.
The second family forgoes a reading and treats the states of a layer as a point cloud, reporting its rank, intrinsic dimension, anisotropy or similarity to other layers \citep{kornblith2019cka,ansuini2019intrinsic,garrido2023rankme,skean2025layer}.
Such statistics describe how the states are distributed rather than how the model uses them. Two states close in the cloud may lead to different predictions, and two distant ones to the same.

Both families omit the frame in which the model itself reads a state. The directions along which a state changes the logit of a token are a property of the remaining layers. These layers attend to the preceding tokens, so the directions differ between contexts, and no method with one map per layer can follow them, however it is fitted.
In OLMo3-7B (Figure~\ref{fig:rulers}), after layer \MidLayerO{} of \NumLayersO{}, the logit-lens direction for a token deviates by \AngLLMidO{} degrees from the model's own lexical readout and the Jacobian-lens direction by \AngJLMidO{} degrees. Four layers before the output, both still deviate by at least \AngLensLateMinO{} degrees.

Our starting point is that these directions, the lexical readouts, form a reference frame. We call their family at a state its \emph{lexical reference frame}, and the frame that the complete model induces its \emph{native frame}.
The reading of the native frame is the model's output at every layer, so an informative frame withholds a chosen part of this computation. The frame of order $n$, written \lrfn{$n$}, transports a state through the model's own layers with $n$ rounds of attention between positions. It replaces the remaining attention by a typical attention output, estimated once per layer from unlabelled text.
\lrfn{0} reads what a state holds by itself, \lrfn{1} what it draws from its prefix, and \lrfn{$\infty$} is the native frame.
The idea is to read a state the way the model reads it, through the layers that remain and with the context it sees. Such a frame is model-native and context-aware up to its order, while a lens ignores the context and is therefore bounded.
As nothing is fitted to the model's output, the frames are comparable across layers, checkpoints and models of a family.

We study lexical reference frames in two recent open-weight models, SmolLM3-3B \citep{smollm3} and OLMo3-7B \citep{olmo2025olmo3}. We compare them with lenses, with statistics of the point cloud and with attribution patching.
From mid-depth on, \lrf{} recovers the native frame's lexical readouts beyond the bound of every lens. It matches the model's output and its calibration, which lenses do only near the output. One round of attention to the prefix closes most of the gap that \lrfn{0} leaves.
Its readings settle on the model's top token from mid-depth on at most positions, whereas those of lenses settle only near the output. In one example, \lrf{} already names the model's answer at layer \ExOneLRF{}, where the Jacobian lens does not (Appendix~\ref{app:more}). Over training, its readings remain faithful while the logit lens degrades.
Distances on the point cloud poorly predict the effect of replacing a state, as lexical readouts lie almost entirely outside the directions in which the states vary.
In the coordinates of the frame, the lexical basis writes a prediction into the stream more selectively than the readout. Differences of coefficients attribute it to layers and sublayers more faithfully than attribution patching.
Lexical reference frames can therefore stand in for lenses wherever intermediate states are read.
We make the following three contributions.

\begin{itemize}
\item We identify the native reference frame of a hidden state, the frame induced by the layers that remain in its context, and prove that no lens with fixed directions per layer can follow it beyond a bound set by its coherence.
\item We define lexical reference frames, which read a state through the remaining layers with a given number of rounds of attention between positions. They are calibrated in closed form and fit nothing to the model's output.
\item We show on two model families and their training checkpoints that these frames follow the native frame from mid-depth on, where lenses fail, and that their basis and coefficients steer and attribute predictions at every layer.
\end{itemize}

\section{Related Work}

Prior work on the intermediate layers of language models asks which tokens a state stands for, how the states of a layer are distributed and which directions or components drive a prediction.

\paragraph{Language model lenses.}
The logit lens reads an intermediate state with the model's own output frame \citep{nostalgebraist2020logitlens}, and it underlies studies of how predictions are refined across depth \citep{jastrzebski2017residual,geva2020transformer,geva2022concepts,dar2023embedding,halawi2023overthinking,katz2024backward,wendler2024llamas,lad2024remarkable}.
Later lenses use a map per layer, trained to match the output \citep{belrose2023tuned,din2023jump} or given by a Jacobian of the remaining layers \citep{hernandez2023linearity,gurnee2026workspace}, or patch the state into another forward pass \citep{ghandeharioun2024patchscopes,pal2023future}.
\citet{gu2026semrf} fix a set of anchors and measure the states of all layers against them. \citet{gu2026tying} tie the input and output frames to each other through the pseudo-inverse of one shared token memory.
Lenses are thus frames that ignore the context, and we bound how closely any can approximate the native frame.

\paragraph{Point clouds and manifolds.}
A large body of work summarises the states of a layer by statistics of the cloud they form, such as similarity between layers \citep{raghu2017svcca,kornblith2019cka}, intrinsic dimension \citep{facco2017twonn,ansuini2019intrinsic,valeriani2023geometry}, effective rank \citep{garrido2023rankme} and anisotropy \citep{ethayarajh2019contextual,gao2019representation,razzhigaev2023shape,bernas2026anisotropy}.
Recent studies follow these statistics across layers \citep{cheng2024emergence,skean2025layer} and across training \citep{saphra2018understanding,tigges2024llm,lee2025geometric}.
We quantify how much these statistics reveal about the way the model uses the states of a layer.

\paragraph{Directions and components.}
Linear probes learn a direction from labelled states \citep{alain2017probes,tenney2019bert,hewitt2019designing,belinkov2021probing}, and steering methods add a direction to the states to change the output \citep{subramani2022extracting,turner2023steering,li2023inference,zou2023representation,rimsky2024caa,arditi2024refusal}.
Both rest on the hypothesis that concepts are represented as directions \citep{park2023linear}, which \citet{park2026softmax} derive from the information geometry of the softmax for the output layer only.
The contribution of a component to a prediction is likewise computed along a direction, by projecting its write onto the output frame \citep{wang2022interpretability}, or measured by patching \citep{meng2022locating,conmy2023automated,syed2023attribution,kramar2024atp}.
A frame supplies such directions for each state without supervision, which we test in steering and attribution at every layer.

\section{Lexical Reference Frames}
\label{sec:method}

We describe a hidden state by the frame in which the model reads it in its vocabulary. It is fixed at both ends of the stream, adaptive in between and approximated at finite order (Appendix~\ref{app:theory}).

\begin{figure*}[htb]
\centering
\includegraphics[width=\textwidth]{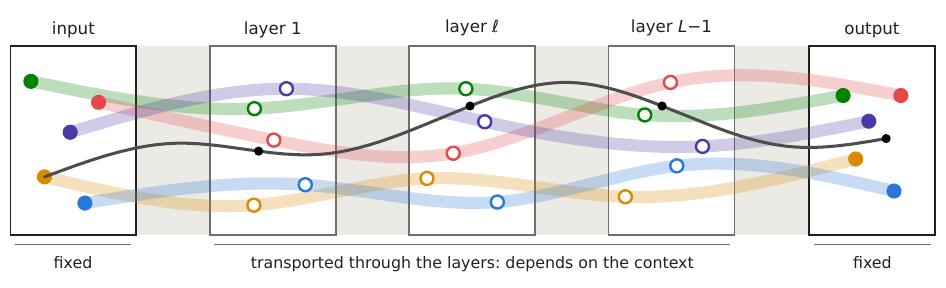}
\caption{Lexical bases of five tokens (colours) and one state (black) across the layers, fixed at the two interfaces (filled) and shifting from layer to layer in between (hollow).}
\label{fig:transport}
\end{figure*}

\subsection{Preliminaries}
\label{sec:prelim}

Layers are numbered from 1 to $L$, and layer 0 denotes the embedding.
Let $x_\ell(t) \in \mathbb{R}^d$ be the hidden state, the residual stream after layer $\ell$ at position $t$.
Layer $k$ reads the stream and writes its update back into it \citep{elhage2021mathematical,veit2016residual}, $x_{k}(t) = F_k\bigl(x_{k-1}(t),\, a_k(t)\bigr)$. Here $a_k(t)$ is the output of its attention at position $t$, which depends on the states of the prefix. $F_k$ contains the residual connections, the normalisations and the feed-forward block.
The native semantics of the model are its vocabulary labels, the smallest units of meaning that it receives and produces, in which a reference frame expresses a state.

At its two interfaces the model itself gives every label a geometric form.
At the input, the embedding assigns label $v$ the state $E_v$.
At the output, the logit of $v$ is $z_v = w_v^\top N(x_L(t))$ with the final normalisation $N$ and the row $w_v$ of the unembedding matrix $W$, and its centred value is $c_v = z_v - \frac{1}{|V|}\sum_{u} z_u$.
The normalised state whose logits come closest, in least squares, to the indicator $e_v$ of $v$ is $s_v = W^{+} e_v$, which \citet{gu2024vocabulary} call its semantic basis.
We call such a family of vectors, one per label, a \emph{lexical basis}.
Weighted by the logits, the output lexical basis sums to $W^{+} W N(x_L(t)) = N(x_L(t))$, so at the output a normalised state is composed from its labels and decomposed into them exactly.
The input lexical basis and the output lexical basis are the same for every text, so the reference frames that they span, the input frame and the output frame, are \emph{fixed}.
Earlier work measures states against these fixed frames and decomposes them into coefficients on the output lexical basis, to select layers for fine-tuning \citep{gu2025freezing} and to transfer knowledge across model scales \citep{gu2025alignment}.

\subsection{Native Reference Frame}
\label{sec:native}

Between the interfaces the model defines no basis, and a frame must be \emph{adaptive}, native to the model and aware of the context.
A state receives its meaning from the layers that remain, which read it from the stream.
They transport $x_\ell(t)$, together with the states of its prefix, to the last layer, where the output frame applies.
We call this passage \emph{native transport} and write $T_\ell$ for the map from $x_\ell(t)$ to $x_L(t)$ with the states of the prefix held fixed.
The output frame reaches layer $\ell$ by being pulled back through this map,
\begin{equation}
\label{eq:frame}
b_\ell(v) = \frac{\partial c_v}{\partial x_\ell(t)} = J_\ell^\top (w_v - \bar w) ,
\end{equation}
where $J_\ell = \partial N(T_\ell) / \partial x_\ell(t)$ is the Jacobian of the transport followed by the final normalisation, and $\bar w$ is the mean unembedding row.
We call $b_\ell(v)$ the \emph{lexical readout} of label $v$ at the state, and the family of these directions over the vocabulary is the \emph{lexical reference frame} of the state.
The lexical basis at the layer is $s_\ell(v) = J_\ell^{+} s_v$, the smallest move of the state whose first-order effect on the output comes closest to $s_v$.
Both are determined anew for every state, as Figure~\ref{fig:transport} illustrates. They depend on the layer, since $J_\ell$ contains all remaining layers, and on the context, since these layers attend to the prefix.
The coefficient of a state on label $v$ is the inner product that the model itself takes, applied after the transport,
\begin{equation}
\label{eq:coefficient}
z_\ell(v) = \bigl\langle\, w_v ,\; N\bigl(T_\ell(x_\ell(t))\bigr) \bigr\rangle ,
\end{equation}
and the \emph{reading} of the state is the distribution over the next token that these coefficients give.
The frame that the complete model induces is the \emph{native reference frame} of the state, or its native frame for short.
Basis and coefficients give the stream coordinates in the vocabulary at every layer. The coefficients read the stream and the basis writes into it, while differences of coefficients attribute a prediction to the layers and sublayers that write it.

The dependence on the context is quantified by the coherence $\kappa_\ell(v)$, the mean cosine between the unit lexical readouts of label $v$ at states from different sequences, and $\kappa_\ell$ is its mean over labels.
A lens is specific to the layer but not to the context, as it offers one direction per label and layer.
The mean cosine of a fixed unit vector $r$ with unit lexical readouts $\hat b_i$ is $r^\top \bar b \leq \lVert \bar b \rVert$, where $\bar b$ is their mean.
Since $\lVert \bar b \rVert^2$ tends to $\kappa_\ell(v)$ over many states, no lens exceeds a mean cosine of $\sqrt{\kappa_\ell(v)}$ with the native frame, and the mean direction attains it. Every analysis that reads directions through a lens, such as the vocabulary labels of features, direct logit attribution or steering along a lens direction, inherits this bound.
The native frame is not a ground truth to be recovered but a reference against which any interpretation of a state, and any intervention on it, can be checked. Its reading, however, is the model's output. An informative reading withholds part of the remaining computation.

\subsection{Orders of Reference Frames}
\label{sec:orders}

\begin{figure*}[htb]
\centering
\includegraphics[width=\textwidth]{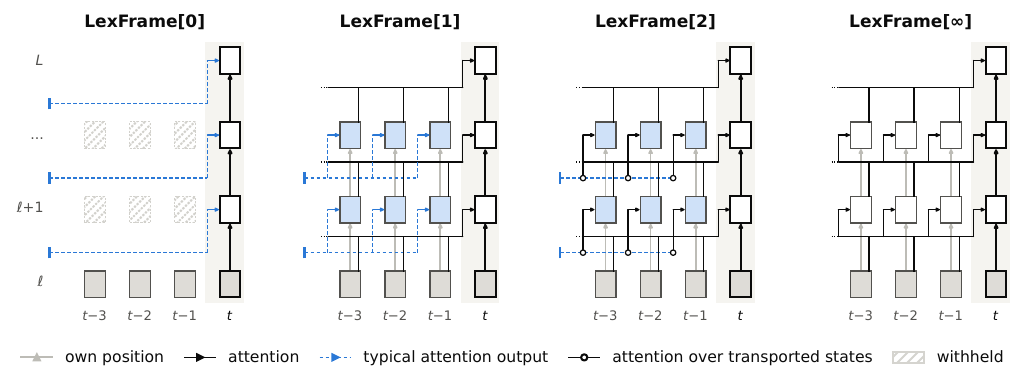}
\caption{What a frame of each order computes to read the state at position $t$ of layer $\ell$, with the layer's own attention (solid black) and typical attention outputs (dashed blue). The order in brackets counts the rounds of attention between positions above layer $\ell$. The context reaches the state only through the state itself (0), directly from the prefix (1), through one other prefix position (2) or without limit ($\infty$).}
\label{fig:orders}
\end{figure*}

A frame of order $n$ replaces the transport $T_\ell$ by one that computes only $n$ rounds of attention between positions. It therefore has its own lexical readouts, lexical basis and reading.
Information can thus pass between positions at most $n$ times above layer $\ell$.
A frame that withholds more is further from the model but more informative about the state. As nothing in it is fitted to the model's output, its agreement with the model is not built in.
As Figure~\ref{fig:orders} shows, every remaining layer updates the state at position $t$ from the state one layer below and its attention over the prefix. All frames keep the first input and differ only in the second.

\paragraph{Order 0.}
\lrfn{0} withholds the prefix and reads a state without its context, and its reading is what the state holds by itself.
It transports $x_\ell(t)$ through the layers $k > \ell$ with the model's own $F_k$, and it replaces the attention output by a typical one,
\begin{equation}
\label{eq:typical}
\hat a_k(y) = \bar a_k + G_k\, (y - \bar y_k) ,
\end{equation}
where $y$ is the attention input at position $t$, and $\bar y_k$ and $\bar a_k$ are the means of the attention input and output of layer $k$. $G_k$ is the ridge regression of the attention output on its input, fitted once for each layer on unlabelled calibration text.
Since its transport keeps every remaining layer, \lrfn{0} still reads a state through them, unlike a lens with one map per layer. It lacks only what the remaining layers would gather from other positions.

\paragraph{Order 1.}
\lrfn{1} lets the state attend to its prefix.
The state at position $t$ is transported as before, but in every layer its attention is the layer's own attention, computed over the prefix.
The states of the prefix are themselves transported from layer $\ell$ by \lrfn{0}, so the frame withholds only the attention among the prefix positions after layer $\ell$.
The reading is what the state draws from its prefix, and the step from \lrfn{0} measures what this one round of attention contributes.

\paragraph{Order 2.}
In \lrfn{2} the attention also reaches within the prefix, since every prefix position attends with the layer's own attention to its own prefix, transported by \lrfn{0}.
Information can therefore travel two steps, from an earlier token through a prefix position to the state, a route that \lrf{} cannot represent.
The step from \lrfn{1} to \lrfn{2} measures what this exchange among the prefix positions contributes.
In general \lrfn{$n$} attends to prefix states of order $n-1$, so each order adds one round of attention. Reading at order $n$ therefore runs $n+1$ streams through the remaining layers.

\paragraph{Order $\infty$.}
In the limit nothing is withheld, and \lrfn{$\infty$} recovers the model as the native frame. Every position then attends to the true states of its prefix, the states that the model itself computes.
Two identities follow. The reading of every layer in \lrfn{$\infty$} equals the model's output, and \lrfn{$n$} is exact wherever at most $n$ layers remain.
Near the output all orders therefore come close to the model. They differ where much of the computation lies ahead.
The value of \lrfn{$\infty$} lies not in its reading but in its lexical readouts and basis, which frames of finite order approximate.

\begin{figure*}[htb]
\centering
\includegraphics[width=\textwidth]{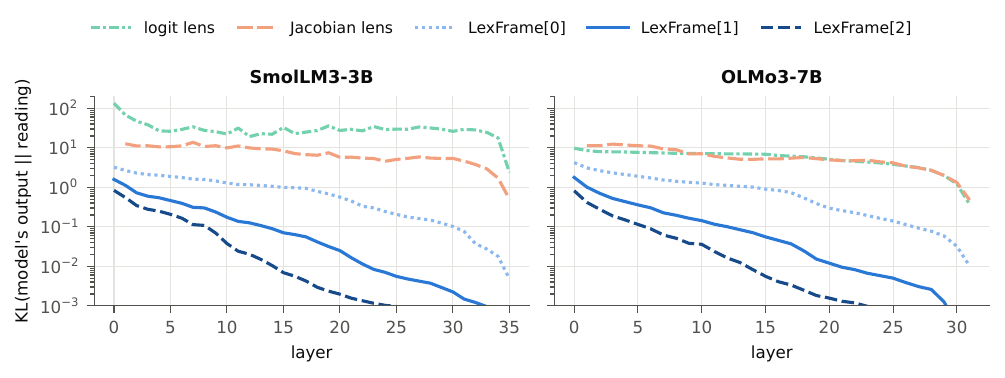}
\caption{Divergence in nats from the model's output to the reading of each frame and each lens, per layer. Each further order lowers it at nearly every layer, while both lenses stay several nats above all frames until near the output.}
\label{fig:readings}
\end{figure*}

\section{Experiments}

We compare frames with lenses and point clouds, then test their coordinates and their use in steering and attribution.
Appendix~\ref{app:setup} details the setup, and Appendix~\ref{app:more} adds further results.

\paragraph{Setup.}
We use SmolLM3-3B with \NumLayersS{} layers and OLMo3-7B with \NumLayersO{} layers, recent competitive open-weight models that release checkpoints from every stage of training. SmolLM3 ties its input and output embeddings, and OLMo3 does not.
We run them in the precision of the released weights and report pairs of values in this order.
The text is taken from WikiText-103, cut into sequences of \SeqLen{} tokens.
The typical attention outputs are estimated on \CalibSeqs{} sequences of the training split, and every measurement uses the test split.
The Jacobian lens is fitted with its reference implementation on \JLPromptsO{} prompts.
We omit the tuned lens, which is trained on the model's output, the quantity on which readings are scored, and is bounded in direction like every lens.
As a supervised reference we add a linear probe, a ridge regression of the centred logit on the states, fitted on separate text.
Divergences are Kullback--Leibler divergences from the model's output to a reading, in nats, and averages over layers are unweighted.
\lrfn{$\infty$} is the reference for readouts and the exact frame in interventions.

\subsection{Comparison with Lenses}
\label{sec:fidelity}

We compare frames and lenses in the directions along which they measure a label and in the distributions that they assign to a state.
At intermediate layers, \lrfn{$\infty$} departs sharply from the output frame, and its lexical readouts vary strongly between contexts.
We compare each method's direction for a label, the gradient of its reading or its fixed vector, with these readouts for \DirTokensO{} labels across the frequency ranking.
\lrf{} follows them closely from mid-depth on, because its transport runs through the same layers and withholds only the attention among the prefix positions.
The lenses and the probe remain far below. Their failure is structural rather than a matter of fitting, as no single direction can match a readout that changes with the context.
At mid-depth the bound $\sqrt{\kappa_\ell}$ is \SqrtKappaMidS{} and \SqrtKappaMidO{}, whereas \lrf{} reaches \CosLRFMidS{} and \CosLRFMidO{}. \lrfn{2} comes closer still by adding one round of the exchange among prefix positions that \lrf{} withholds in every remaining layer.

The divergence from the model's output (Figure~\ref{fig:readings}) shows the same ranking, and the first round of attention closes most of the divergence of \lrfn{0}.
From mid-depth on, a state and one round of attention to its prefix determine nearly all of its reading. Each further round of attention lowers the divergence, so the order makes the use of the context a controlled variable.
The lenses diverge by several nats, whereas \lrfn{0}, which uses no more of the context than they do, stays below one nat on average. Transport through the model's own layers hence matters more than a fitted map.
The gap between lenses and frames is wider in SmolLM3, whose output frame is also its input frame. At the embedding its logit lens returns the input token itself at most positions, whereas that of OLMo3, with separate input and output embeddings, almost never does so.

\begin{table}[htb]
\centering
\setlength{\tabcolsep}{4pt}
\begin{tabular}{lrrr}
\toprule
\textbf{Method} & \textbf{KL} $\downarrow$ & \textbf{Top-1} $\uparrow$ & \textbf{ECE} $\downarrow$ \\
\midrule
\multicolumn{4}{l}{\textit{SmolLM3-3B}} \\
Logit lens & 29.06 & 0.04 & 0.632 \\
Jacobian lens & 7.51 & 0.14 & 0.376 \\
\lrfn{0} & 0.87 & 0.64 & 0.018 \\
\lrfn{1} & 0.16 & 0.86 & 0.027 \\
\lrfn{2} & 0.06 & 0.92 & 0.020 \\
\midrule
\multicolumn{4}{l}{\textit{OLMo3-7B}} \\
Logit lens & 5.66 & 0.15 & 0.107 \\
Jacobian lens & 6.19 & 0.13 & 0.191 \\
\lrfn{0} & 0.93 & 0.64 & 0.045 \\
\lrfn{1} & 0.15 & 0.86 & 0.018 \\
\lrfn{2} & 0.05 & 0.93 & 0.020 \\
\bottomrule
\end{tabular}

\caption{Layer-averaged readings: divergence from the model, top-token agreement and calibration error.}
\label{tab:readings}
\end{table}

In Table~\ref{tab:readings}, averaged over layers, \lrfn{0} diverges from the model's output by \KLZeroS{} and \KLZeroO{} nats, \lrf{} by \KLLRFS{} and \KLLRFO{}, and \lrfn{2} by \KLTwoS{} and \KLTwoO{}.
\lrf{} agrees with the model's top token on \TopLRFS{} and \TopLRFO{} of the states, and its agreement exceeds nine tenths from layer \AgreeNineLRFS{} and \AgreeNineLRFO{} on. The lenses reach an agreement of one half only from layer \AgreeHalfLensS{} and \AgreeHalfLensO{} on.
\lrf{} does not merely reproduce the output. At the embedding, where all computation remains, it agrees on only \TopLRFFirstS{} and \TopLRFFirstO{}.
Its calibration error against the actual next token, \ECELRFS{} and \ECELRFO{}, is close to that of the model itself, \ECEModelS{} and \ECEModelO{}. Its probabilities can therefore be interpreted as a prediction in their own right, unlike those of the lenses.
Appendix~\ref{app:more} reports how \lrf{} relaxes toward \lrfn{$\infty$} across depth, from which depth the readings settle, what an early exit can skip and how the readings change over training.

\subsection{Comparison with Point-Cloud Geometry}
\label{sec:cloud}

Without a frame, a point cloud relates to the model's computation only through distances.
\begin{figure}[htb]
\centering
\includegraphics[width=\columnwidth]{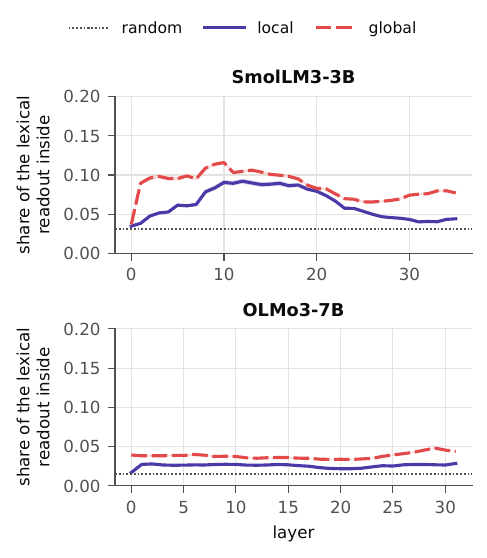}
\caption{Share of a lexical readout's squared norm in 64 local or global principal directions of the cloud.}
\label{fig:tangent}
\end{figure}
We replace the state of one position by the real state of another position at the same layer, complete the forward pass and measure the change in the output, for \SubPairsS{} pairs per layer.
In Table~\ref{tab:substitution}, averaged over layers, no distance on the cloud exceeds a rank correlation of \SubCloudHi{} with this change, and the geodesic distance, which follows the curvature of the cloud, performs worse than the plain cosine distance.
In \lrf{}, the divergence between the readings of the two states correlates with this change at \SubLRFS{} and \SubLRFO{}, and already \SubLRFFirstS{} and \SubLRFFirstO{} at the embedding. The frame compares the two states in the directions to which the remaining layers respond.
A lens, which reads along fixed directions, reaches a rank correlation of at most \SubJLS{}, so a reading must follow the frame as well.

\begin{table}[htb]
\centering
\setlength{\tabcolsep}{2.5pt}
\begin{tabular}{lrr}
\toprule
\textbf{Distance} & \textbf{SmolLM3-3B} & \textbf{OLMo3-7B} \\
\midrule
Logit lens & 0.11 & 0.16 \\
Jacobian lens & 0.28 & 0.24 \\
\midrule
Cosine & 0.27 & 0.28 \\
Euclidean & 0.08 & 0.12 \\
Principal subspace & 0.09 & 0.13 \\
Mahalanobis & 0.08 & 0.03 \\
Geodesic & 0.09 & 0.07 \\
\midrule
\lrfn{1} & 0.94 & 0.95 \\
\bottomrule
\end{tabular}

\caption{Layer-averaged rank correlation between a distance of two states and the effect of their substitution.}
\label{tab:substitution}
\end{table}

Figure~\ref{fig:tangent} explains this gap. For each state we estimate the tangent space of the cloud from its 128 nearest states, or from the whole cloud, and measure the share of a lexical readout in its 64 leading directions.
This share is small, at most a few times that of a random direction.
As a lexical readout is the gradient of a logit, moves along the directions in which the cloud varies change each logit far less than equal moves along the readout. Distances on the cloud thus mainly measure variation to which the remaining layers do not respond. A point cloud thus describes how the states of a layer vary, whereas a frame describes how the model itself reads each of these states.

\subsection{Coordinates at Intermediate Layers}
\label{sec:bases}

At the output, the vocabulary is a coordinate system for the residual stream. The lexical basis gives the axes, the directions in which writing a label moves a state, and the coefficients give the coordinates, how much of each label the state holds.
Weighted by the coefficients, the basis composes the state, and the inner products decompose it into the logits, both exactly.
A matching readout alone therefore does not make a frame at an intermediate layer. The frame must also keep both properties. We test composition, whether a move along the lexical basis of a label composes that label, and decomposition, whether the coefficients of a frame decompose the state into the logits of the model.

\begin{table}[htb]
\centering
\setlength{\tabcolsep}{1.5pt}
\begin{tabular}{lrr}
\toprule
\textbf{Direction} & \textbf{SmolLM3-3B} & \textbf{OLMo3-7B} \\
\midrule
Input frame & 0.42 & 0.00 \\
Output frame & 0.45 & 0.80 \\
\midrule
\lrfn{1}, {\small readout} & 0.54 & 0.73 \\
\lrfn{1}, {\small basis} & 0.84 & 0.97 \\
\lrfn{$\infty$}, {\small readout} & 0.56 & 0.79 \\
\lrfn{$\infty$}, {\small basis} & 0.98 & 1.00 \\
\bottomrule
\end{tabular}

\caption{Share of moves along the lexical readout or basis of a label that raise its logit most, averaged over layers, where fixed frames use their basis unchanged.}
\label{tab:basis}
\end{table}

For composition, we compute $J_\ell$ in 32-bit precision for \BasisStatesS{} states per layer and move each state by \BasisStep{} of its norm along $s_\ell(v)$ for \BasisLabelsS{} labels. Table~\ref{tab:basis} counts how often $v$ is then the label whose logit gains most.
For the lexical basis at the layer this holds in \BasisPulledS{} and \BasisPulledO{} of the cases, from the first layers on.
The lexical readout of \lrfn{$\infty$} succeeds less often, in \BasisReadoutS{} and \BasisReadoutO{}. As the gradient $J_\ell^\top (w_v - \bar w)$ of the centred logit of $v$, it raises $v$ fastest but also every label whose logit moves with that of $v$. It agrees in direction with $J_\ell^{+} s_v$ only where the remaining layers and the unembedding both stretch all directions alike.
The direction that measures a label is therefore not the direction that moves the output toward it.
Fixed frames perform worse than the basis. The output frame, with $s_v$ unchanged, reaches \BasisUnchangedS{} and \BasisUnchangedO{}. The input frame reaches \BasisInputS{} in SmolLM3, which ties its input and output embeddings, and \BasisInputO{} in OLMo3, which does not.
With the Jacobian of \lrf{}, the share is \BasisOrdOneS{} and \BasisOrdOneO{}, against \BasisReadoutOrdOneS{} and \BasisReadoutOrdOneO{} for its readout, as in \lrfn{$\infty$}.

For decomposition, Table~\ref{tab:composition} tests whether the coefficient of a label, the inner product of its unembedding row with the transported state, can do without the transport.
With the state of the layer itself, or with its image under one fixed map per layer, it is several nats from the model's output. With the state that \lrf{} transports, it is within \CmpIPFrameKLS{} and \CmpIPFrameKLO{} nats, and with the transport of \lrfn{$\infty$} it equals the model's logits.
Nor can the transport be linearised. The inner product of the lexical readout of \lrfn{$\infty$} with the state of its layer has a correlation of only \CmpPairModelS{} and \CmpPairModelO{} with the centred logit. The readout is the slope of the logit at the state and not a measure of how much of the label the state holds.
An adaptive frame is local. Basis and coefficients hold at one state and are tied by the transport, not by one matrix as at the output, so composition and decomposition reach intermediate layers only through the transport.

\begin{table}[htb]
\centering
\setlength{\tabcolsep}{2.5pt}
\begin{tabular}{lrrrr}
\toprule
 & \multicolumn{2}{c}{\textbf{SmolLM3-3B}} & \multicolumn{2}{c}{\textbf{OLMo3-7B}} \\
\cmidrule(lr){2-3}\cmidrule(lr){4-5}
\textbf{Paired state} & \textbf{KL} $\downarrow$ & \textbf{Top-1} $\uparrow$ & \textbf{KL} $\downarrow$ & \textbf{Top-1} $\uparrow$ \\
\midrule
Layer state & 8.36 & 0.04 & 5.64 & 0.15 \\
Fixed map & 5.19 & 0.13 & 5.46 & 0.13 \\
\lrfn{1} & 0.16 & 0.87 & 0.15 & 0.84 \\
\lrfn{$\infty$} & 0.00 & 1.00 & 0.00 & 1.00 \\
\bottomrule
\end{tabular}

\caption{Coefficients on the output lexical basis against model logits on \CmpRowsS{} positions, averaged over layers: divergence (nats) at the best scale, top-token agreement.}
\label{tab:composition}
\end{table}

\subsection{Steering and Attribution}
\label{sec:uses}

The two parts of these coordinates serve two interventions on the stream. Steering writes into it along the basis, and attribution reads through the coefficients what a component wrote.
In steering, the task is to make the runner-up token $u$, the target that needs the smallest move, overtake the top token $v$ by moving a single state.
A frame offers two directions for this: the lexical readout of the logit difference, $b_\ell(u) - b_\ell(v)$, which is its gradient, and the difference of the lexical basis vectors, $s_\ell(u) - s_\ell(v)$.
We move each state along each direction in steps from 0.01 to 1 times its norm. We record the first step that flips the prediction and the change in the other tokens at that step.

Along the lexical readout (Table~\ref{tab:steering}), \lrf{} flips \SteerLRFReachS{} and \SteerLRFReachO{} of the predictions, nearly as many as \lrfn{$\infty$} with \SteerNativeReachS{} and \SteerNativeReachO{}, and the lenses at most \SteerLLReachO{}.
The lexical basis of \lrfn{$\infty$} flips nearly as many predictions, but it disturbs the other tokens less, \SteerBasisSideS{} and \SteerBasisSideO{} nats at the flip against \SteerNativeSideS{} and \SteerNativeSideO{}. It therefore yields the largest share of clean flips, \SteerBasisSelS{} and \SteerBasisSelO{} of the predictions.
The readout is the steepest ascent of the logit difference and spreads the change over every token moving with $u$ or $v$. The basis is the move whose first-order effect on the output comes closest to raising $u$ against $v$ alone.
The basis computed from the Jacobian of \lrf{} still yields clean flips for \SteerBasisLRFSelS{} and \SteerBasisLRFSelO{} of the predictions. Selective steering thus works even without access to the Jacobian of the model itself.

\begin{table}[htb]
\centering
\setlength{\tabcolsep}{2.5pt}
\begin{tabular}{lrrr}
\toprule
\textbf{Direction} & \textbf{Reach} $\uparrow$ & \textbf{Side} $\downarrow$ & \textbf{Clean} $\uparrow$ \\
\midrule
\multicolumn{4}{l}{\textit{SmolLM3-3B}} \\
Logit lens & 0.69 & 0.20 & 0.49 \\
Jacobian lens & 0.78 & 0.37 & 0.49 \\
\lrfn{1}, {\small readout} & 0.92 & 0.23 & 0.61 \\
\lrfn{1}, {\small basis} & 0.89 & 0.12 & 0.71 \\
\lrfn{$\infty$}, {\small readout} & 0.95 & 0.22 & 0.63 \\
\lrfn{$\infty$}, {\small basis} & 0.90 & 0.11 & 0.73 \\
\midrule
\multicolumn{4}{l}{\textit{OLMo3-7B}} \\
Logit lens & 0.81 & 0.22 & 0.57 \\
Jacobian lens & 0.79 & 0.32 & 0.51 \\
\lrfn{1}, {\small readout} & 0.94 & 0.20 & 0.64 \\
\lrfn{1}, {\small basis} & 0.95 & 0.07 & 0.82 \\
\lrfn{$\infty$}, {\small readout} & 0.97 & 0.19 & 0.67 \\
\lrfn{$\infty$}, {\small basis} & 0.97 & 0.06 & 0.85 \\
\bottomrule
\end{tabular}

\caption{Steering along the lexical readout or basis, averaged over layers: share of predictions flipped, mean change in the other tokens at the first flip (nats) and share flipped with a change below 0.1 nats.}
\label{tab:steering}
\end{table}

In attribution, the task is to quantify how much a component, a layer or one of its two sublayers, contributed to the centred logit of the predicted token.
As the reference we remove what the component wrote at one position, leave the prefix untouched, complete the forward pass and measure the change in the centred logit, on \AttrSeqsS{} sequences.
This effect is the difference of coefficients in \lrfn{$\infty$}, so the test asks whether a frame that withholds exchanges within the prefix still attributes as the model does.
In a frame, the attribution is the change in the coefficient of the predicted token when the write is removed from the state.
We compare it with the same difference in the two lenses, of which the logit lens gives direct logit attribution. We also compare it with linear attributions, the inner product of a lexical readout with the write, which for \lrfn{$\infty$} are attribution patching.

\begin{table}[htb]
\centering
\setlength{\tabcolsep}{3pt}
\begin{tabular}{lrrr}
\toprule
\textbf{Attribution} & \textbf{Layer} & \textbf{Attention} & \textbf{MLP} \\
\midrule
\multicolumn{4}{l}{\textit{SmolLM3-3B}} \\
Logit lens & 0.08 & 0.13 & 0.04 \\
Jacobian lens & 0.12 & 0.18 & 0.10 \\
\lrfn{1}, {\small diff} & 0.88 & 0.79 & 0.88 \\
\lrfn{1}, {\small linear} & 0.57 & 0.67 & 0.58 \\
\lrfn{$\infty$}, {\small linear} & 0.62 & 0.75 & 0.63 \\
\midrule
\multicolumn{4}{l}{\textit{OLMo3-7B}} \\
Logit lens & 0.14 & 0.18 & 0.12 \\
Jacobian lens & 0.15 & 0.20 & 0.12 \\
\lrfn{1}, {\small diff} & 0.90 & 0.85 & 0.90 \\
\lrfn{1}, {\small linear} & 0.62 & 0.75 & 0.64 \\
\lrfn{$\infty$}, {\small linear} & 0.68 & 0.84 & 0.71 \\
\bottomrule
\end{tabular}

\caption{Layer-averaged correlation of attributed and measured effects on the predicted centred logit.}
\label{tab:attribution}
\end{table}

In Table~\ref{tab:attribution}, the difference in \lrf{} correlates with the measured effect at \AttrLRFS{} and \AttrLRFO{}. This is closer than the linear attribution of \lrfn{$\infty$}, at \AttrGradS{} and \AttrGradO{}, although the latter uses the model's Jacobian.
A linear attribution assumes that the effect of a write is proportional to it. Its deficit relative to the difference is large for the feed-forward block (MLP), \AttrMlpLRFS{} and \AttrMlpLRFO{} against \AttrMlpGradS{} and \AttrMlpGradO{}, but small for attention, \AttrAttLRFS{} and \AttrAttLRFO{} against \AttrAttGradS{} and \AttrAttGradO{}. The remaining layers respond almost linearly to attention writes and far beyond first order to feed-forward writes. A difference of coefficients needs no linearisation, so it stays faithful where the response is nonlinear.
The two lenses reach at most \AttrAttJLO{}. The logit lens measures only the direct path from a write to the output, and this direct path accounts for only a small part of the measured effect.

\section{Conclusion}

Every hidden state of a language model has a native reference frame, fixed at the two ends of the residual stream and adaptive in between. Lexical reference frames, which are model-native and context-aware, approximate it up to a chosen order without fitting to the model's output.
In two model families \lrf{} recovers the model's lexical readouts and output from mid-depth on, beyond the reach of any fixed lens, and it attributes and steers predictions.
The frame explains why a fixed lens fails at mid-depth, increasingly so over training, and why the point cloud reveals little about how the model uses a state.
To interpret a state, read it as the model does: through the layers that remain, with the context it sees and in its own native semantics, a reading that can stand in for lenses.

\bibliography{refs}

\appendix

\section{Technical Details}
\label{app:setup}

\subsection{Implementation Details}

\paragraph{Models and data.}
The SmolLM3 checkpoints follow a grid in which the number of training tokens roughly triples from one checkpoint to the next. The OLMo3 checkpoints are chosen at matching token counts, with the randomly initialised model and an early checkpoint at four billion tokens in addition.
Both families include the checkpoints at the end of their later training stages.
Every average over layers runs over layers 1 to $L-1$, without the embedding, where the Jacobian lens is undefined, and the output, where every reading is exact.
We calibrate on \CalibSeqs{} sequences of \SeqLen{} tokens from the training split and evaluate on \ReadSeqsS{} sequences of \SeqLen{} tokens from the test split.

\paragraph{Estimation methods.}
The frames involve no training. The typical attention outputs are fitted in closed form from one forward pass over the calibration text. The ridge of every layer is chosen from a fixed grid by two-fold cross-validation over the calibration sequences.
Outside the composition test, gradients are taken through the released weights in their stored half precision.
We compute the lexical basis $s_\ell(v) = J_\ell^{+} s_v$ as the solution of $\min_\delta \lVert J_\ell \delta - s_v \rVert^2 + \lambda \lVert \delta \rVert^2$ with $\lambda = 10^{-3}\, \lVert J_\ell^\top s_v \rVert^2 / \lVert s_v \rVert^2$, the same relative ridge in both experiments that use it.
In the composition test we form $J_\ell$ explicitly in 32-bit arithmetic, with the stored weights cast exactly, and solve the problem in closed form for all labels at once. Every move is then evaluated in the model itself, whichever Jacobian produced the direction.
In steering, which needs the basis at every position, we solve the problem without forming $J_\ell$ by 30 steps of LSMR \citep{fong2011lsmr} from zero. Each product $J_\ell \delta$ is then one forward-mode pass, and each product $J_\ell^\top \eta$ is one backward pass.

\paragraph{Evaluation methods.}
Positions and tokens are sampled once with fixed seeds, and every experiment except the context-length test uses the same test sequences. All results are from a single run, which is deterministic up to the rounding of the arithmetic.
The identity that the reading of \lrfn{$\infty$} equals the model's output holds up to the rounding of half precision, with a divergence of \GateTrueS{} and \GateTrueO{} nats.
The calibration error is the expected calibration error of \citet{guo2017calibration} with 15 bins.
Intervals are 95\% percentile bootstrap intervals from at least 300 resamples of the test sequences, so that positions of the same sequence stay together.
They appear as bands in the figures. In the tables, which report means, their half-width is at most \CIKLMax\% of a divergence and \CISteerSide{} nats for the change in the other tokens in steering. It is at most \CIShareMax{} for every other value, except in the ablation of fitted terms, which reports point estimates.

\subsection{Artifacts and Computations}

We use SmolLM3-3B \citep{smollm3} and OLMo3-7B \citep{olmo2025olmo3} with their published intermediate checkpoints, both released under the Apache 2.0 licence.
The text is the English WikiText-103 corpus \citep{merity2017pointer}, released under the Creative Commons Attribution-ShareAlike licence, with about 103 million tokens in its training split and 0.25 million in its test split.
Our use of these artifacts is consistent with their intended use, and the text contains no information that we collected or annotated.
The experiments ran on NVIDIA RTX 4080 SUPER GPUs, and every run uses a single GPU.
The whole project used about \GpuHoursTotal{} GPU hours, of which about \GpuHours{} hours produced the final experiments reported in this paper.
The implementation is written in Python 3.12 with PyTorch 2.12 and Transformers 5.16.

\subsection{AI Use Disclosure}

AI assistants supported the writing of this paper and the preparation of its code.
The authors checked all text, code and results and take full responsibility.

\section{Additional Results}
\label{app:more}

\begin{figure}[htb]
\centering
\includegraphics[width=\columnwidth]{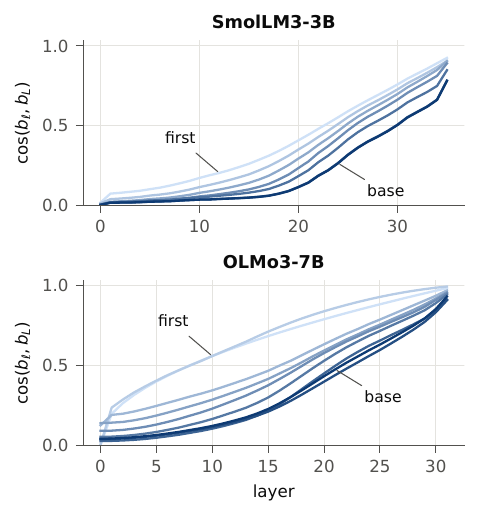}
\caption{Cosine between the lexical readouts of \lrfn{$\infty$} at a layer and at the last layer, from the first checkpoint (lightest) to the base model (darkest).}
\label{fig:formation}
\end{figure}

\paragraph{Training checkpoints.}
\lrfn{$\infty$} is defined through the layers that remain, so it changes with the layer and whenever these layers change.
Figure~\ref{fig:formation} shows this for the labels whose lexical readouts are compared. It plots the cosine between the lexical readout of \lrfn{$\infty$} at a layer and at the last layer of the same forward pass.
In the final models the mid-depth cosine is low, so the output frame is far from the frame that the model uses there.
This departure is learned. At the first checkpoint the mid-depth cosine is much higher, in the randomly initialised OLMo3 and even in SmolLM3, whose earliest released checkpoint is already well into training. It falls over the course of training.
A frame that does not adapt is adequate only in the last layers, where this cosine remains high at every checkpoint of training.

Across checkpoints, the ranking of the readings is stable in both families.
The divergence of the logit lens from the model's output grows from \KLLLTrainFirstS{} and \KLLLTrainFirstO{} to \KLLLTrainLastS{} and \KLLLTrainLastO{} nats, while that of \lrf{} only moves from \KLLRFTrainFirstS{} and \KLLRFTrainFirstO{} to \KLLRFTrainLastS{} and \KLLRFTrainLastO{}.
The decline of the logit lens thus tracks this departure of the frame at mid-depth. A lens degrades not because the model becomes harder to read but because the model's own frames move away from the output frame.
Averaged over layers, \lrf{} agrees with the model's top token on at least \TopLRFTrainLoS{} and \TopLRFTrainLoO{} of the states at every trained checkpoint. The agreement of the logit lens falls from \TopLLTrainFirstS{} to \TopLLTrainLastS{} and from \TopLLTrainFirstO{} to \TopLLTrainLastO{}.
Every value is an average over the layers and over the same \ReadSeqsS{} test sequences, and the typical attention outputs are estimated anew at every checkpoint. The comparison across checkpoints is free of the confound that a lens fitted at one checkpoint is outdated at the next.
The randomly initialised OLMo3 is a control. \lrf{} is within \KLLRFRandO{} nats of its output and the logit lens \KLLLRandO{} nats away, as each frame is built from the layers of its own model.

\begin{figure}[htb]
\centering
\includegraphics[width=\columnwidth]{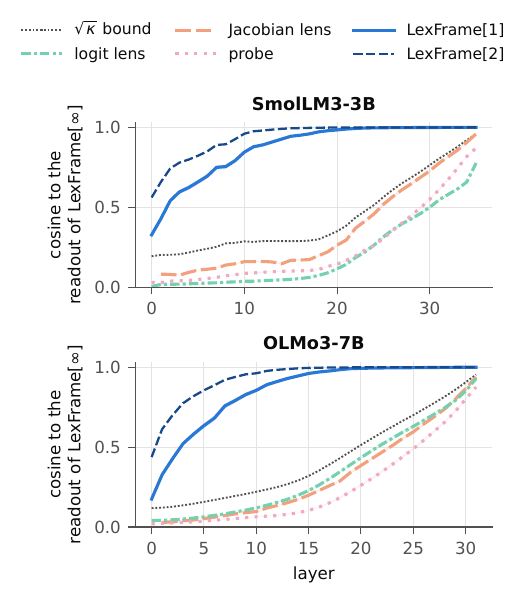}
\caption{Cosine of each method's direction for a label with the lexical readout of \lrfn{$\infty$}, per layer, and the coherence bound $\sqrt{\kappa_\ell}$.}
\label{fig:directions}
\end{figure}

\paragraph{Lexical readouts.}
Figure~\ref{fig:directions} resolves the comparison of lexical readouts by layer.
\lrf{} comes close to \lrfn{$\infty$} from mid-depth on, and \lrfn{2} several layers earlier. The fixed directions approach the frame only in the last layers, where little computation remains.
Even a lens fitted to the model's own Jacobian stays well below the bound $\sqrt{\kappa_\ell}$ at mid-depth, and it never exceeds the bound at any layer.
The bound itself rises toward the output, as the readouts depend less and less on the context, and every fixed direction rises with it.
The frames are weakest at the first layers, where most of the exchange among prefix positions still lies ahead. There \lrfn{2}, which adds one exchange, improves markedly on \lrf{}.

\begin{figure}[htb]
\centering
\includegraphics[width=\columnwidth]{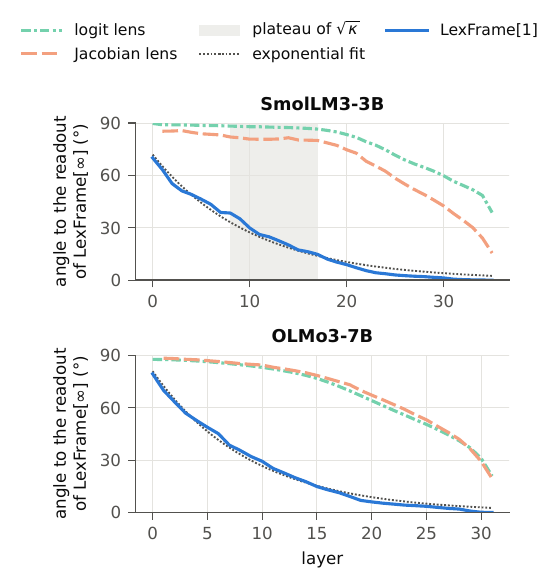}
\caption{Mean angle of each method's direction to the lexical readout of \lrfn{$\infty$} per layer, with the exponential fit for \lrf{} and the plateau in the coherence of SmolLM3 (shaded).}
\label{fig:relax}
\end{figure}

\paragraph{Relaxation across depth.}
The angle between the lexical readouts of \lrf{} and \lrfn{$\infty$} closes at a nearly constant rate across depth (Figure~\ref{fig:relax}). Averaged over states and labels, it shrinks by \RelaxRateS\% and \RelaxRateO\% of what remains at every layer, and one exponential fits all layers with $R^2 = \RelaxRsqS{}$ and \RelaxRsqO{}.
A constant relative decrease per layer is exponential relaxation, the path along which an overdamped system settles toward equilibrium and which the principle of least dissipation selects. The frame thus approaches \lrfn{$\infty$} gradually over the layers that remain, whereas a lens stays far from it until the last layers.
The rate is not exactly constant. In both models the angle lies slightly below the fit in the first layers and from mid-depth on and above it in between, so the relaxation is slower before mid-depth than a constant rate implies. The deviation is larger in SmolLM3, so one exponential fits SmolLM3 less closely than OLMo3. In SmolLM3 the angle of \lrf{} stays above the fit up to layer \RelaxAboveToS{}, whereas in OLMo3 only up to layer \RelaxAboveToO{}, and this slower stretch of \lrf{} coincides with a plateau in the square root of the coherence of \lrfn{$\infty$}, which stays near \PlateauValS{} from layer \PlateauFromS{} to \PlateauToS{}.
The plateau is learned. It is absent at the SmolLM3 checkpoints up to \PlateauLastAbsentS{} trillion training tokens and present from \PlateauFirstS{} trillion tokens on, whereas none of the \CkptsO{} OLMo3 checkpoints shows it. Weight tying and the architecture of SmolLM3 are fixed from the start of training, so neither of them produces the plateau on its own.

\begin{figure}[htb]
\centering
\includegraphics[width=\columnwidth]{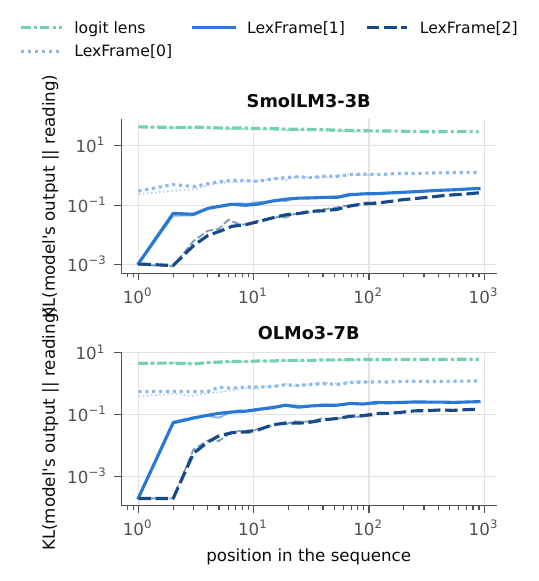}
\caption{Divergence from the model's output to each reading by position, averaged over layers, for sequences of 1024 tokens (thick) and 128 tokens (thin).}
\label{fig:length}
\end{figure}

\paragraph{Context length.}
A longer context offers more positions to attend to and therefore more attention that a frame of finite order withholds, so long sequences test the orders directly.
We repeated the readings on \ReadSeqsS{} sequences of 1024 tokens, with the typical attention outputs estimated on as many calibration tokens as in the main readings.
The divergence of \lrf{} from the model's output rises only from \KLShortLRFS{} and \KLShortLRFO{} to \KLLongLRFS{} and \KLLongLRFO{} nats, and that of \lrfn{2} from \KLShortTwoS{} and \KLShortTwoO{} to \KLLongTwoS{} and \KLLongTwoO{}. The ranking of all methods is unchanged, and \lrf{} still agrees with the model's top token on most states.
The attention among the prefix positions that \lrf{} withholds grows with the context. At intermediate layers, however, it contributes little to what a state means to the model, so a longer prefix adds little to the divergence.
Figure~\ref{fig:length} resolves the divergence by position, that is, by the length of the context. It grows slowly and without a break between the two lengths, so the frame degrades only gradually up to 1024 tokens.

\paragraph{Settling depth.}
Counted back from the last layer, the unbroken run of layers whose reading matches the model's top token shows from which depth a prediction is settled. Early exit relies on this quantity.
Over all \SettleRowsS{} positions (Table~\ref{tab:settle}), \lrf{} is settled over \SettleLRFS{} and \SettleLRFO{} of the layers on average. The Jacobian lens is settled over \SettleJLS{} and \SettleJLO{} and the logit lens over \SettleLLS{} and \SettleLLO{}. Each order lengthens the run, from \SettleZeroS{} and \SettleZeroO{} for \lrfn{0} to \SettleTwoS{} and \SettleTwoO{} for \lrfn{2}.
An exit cannot know the model's top token. It leaves at the first layer whose top-token probability exceeds a threshold.
The threshold is chosen on half of the sequences for 0.95 agreement with the model and applied to the other half. \lrf{} then skips \SettleExitLRFS{} and \SettleExitLRFO{} of the layers, the lenses at most \SettleExitLLO{}. The probabilities of the lenses do not track whether they are right.
\lrf{} saves no computation, as it transports a state through the later layers, but it shows how much depth a lens-based exit forgoes.

\begin{table}[htb]
\centering
\setlength{\tabcolsep}{1.5pt}
\begin{tabular}{lrr}
\toprule
\textbf{Method} & \textbf{Settled layers} $\uparrow$ & \textbf{Skipped layers} $\uparrow$ \\
\midrule
\multicolumn{3}{l}{\textit{SmolLM3-3B}} \\
Logit lens & 0.03 & 0.00 \\
Jacobian lens & 0.09 & 0.00 \\
\lrfn{0} & 0.58 & 0.23 \\
\lrfn{1} & 0.82 & 0.45 \\
\lrfn{2} & 0.89 & 0.53 \\
\midrule
\multicolumn{3}{l}{\textit{OLMo3-7B}} \\
Logit lens & 0.11 & 0.04 \\
Jacobian lens & 0.09 & 0.03 \\
\lrfn{0} & 0.58 & 0.20 \\
\lrfn{1} & 0.81 & 0.42 \\
\lrfn{2} & 0.90 & 0.54 \\
\bottomrule
\end{tabular}

\caption{Mean share of layers, counted back from the last, over which each reading settles on the model's top token, and mean share that a threshold-based exit skips.}
\label{tab:settle}
\end{table}

\begin{table}[htb]
\centering
\setlength{\tabcolsep}{3pt}
\begin{tabular}{lrrr}
\toprule
\textbf{Fitted terms} & \textbf{KL} $\downarrow$ & \textbf{Cosine} $\uparrow$ & \textbf{Attribution} $\uparrow$ \\
\midrule
\multicolumn{4}{l}{\textit{SmolLM3-3B}} \\
Regression & 0.16 & 0.89 & 0.88 \\
Mean & 1.26 & 0.73 & 0.67 \\
Zero & 0.76 & 0.71 & 0.68 \\
\midrule
\multicolumn{4}{l}{\textit{OLMo3-7B}} \\
Regression & 0.15 & 0.87 & 0.90 \\
Mean & 0.52 & 0.74 & 0.77 \\
Zero & 0.47 & 0.74 & 0.77 \\
\bottomrule
\end{tabular}

\caption{\lrf{} with fitted terms (regression), mean or zero output, averaged over layers: divergence (nats), cosine with \lrfn{$\infty$} and attribution.}
\label{tab:typical}
\end{table}

\paragraph{Ablation of fitted terms.}
The fitted terms of a frame are those of the typical attention output, the means $\bar y_k$ and $\bar a_k$ and the ridge regression $G_k$. Table~\ref{tab:typical} tests whether the regression matters by keeping only the mean $\bar a_k$ or by setting the output to zero.
Both replacements make every measure of \lrf{} worse. Its divergence from the model's output grows several times. Both the cosine of its lexical readouts with those of \lrfn{$\infty$} and the correlation of its attribution with the measured effect drop markedly.
The attention that a frame withholds is therefore not a constant. It varies with the input of the attention, and one linear map per layer already captures enough of this variation to keep the frame close to the model.
On three disjoint calibration texts, the divergence lies between \CalLoS{} and \CalHiS{} nats in SmolLM3 and between \CalLoO{} and \CalHiO{} in OLMo3, so the frame depends little on the calibration text.

\paragraph{Worked example.}
Table~\ref{tab:example} shows a position where OLMo3 continues ``\ExContext{}'' with \ExToken{} at probability \ExProb{} rather than \ExRunnerUp{} at \ExProbRunnerUp{}.
Among the listed layers, \lrf{} names this token from layer \ExOneLRF{} on, the Jacobian lens from layer \ExOneJL{} on and the logit lens from layer \ExOneLL{} on. The frame thus reads the prediction well before either lens.
Below that layer \lrf{} already reads whole words that can follow ``The'' here, and at layer \ExRunnerUpLayer{} the runner-up \ExRunnerUp{}. The state thus holds the two newspaper titles that the model weighs at the output, and the next layers settle between them.
The logit lens returns mostly word fragments until layer \ExOneLL{}, because it reads every state along the output directions, which intermediate states do not yet use.
The Jacobian lens returns unrelated tokens until layer \ExOneJL{}, since its map averages over contexts and cannot follow this one, although it is fitted to the model's own Jacobian.
The position shows in a single case what the statistics over all positions report in aggregate. The frame follows the model's own choice, including the alternative it weighs, several layers before either of the two lenses does.

\begin{table}[htb]
\centering
\setlength{\tabcolsep}{2pt}
\begin{tabular}{clll}
\toprule
\textbf{Layer} & \textbf{Logit lens} & \textbf{Jacobian lens} & \textbf{\lrfn{1}} \\
\midrule
0 & ``\tok{oretical}'' & -- & ``\tok{New}'' \\
4 & ``\tok{oretical}'' & ``\tok{flavours}'' & ``\tok{Bill}'' \\
8 & ``\tok{nto}'' & ``\tok{urga}'' & ``\tok{Bill}'' \\
12 & ``\tok{comparison}'' & ``\tok{...\textbackslash{}n\textbackslash{}n}'' & ``\tok{Times}'' \\
16 & ``\tok{orem}'' & ``\tok{Illustrated}'' & ``\tok{Guardian}'' \\
20 & ``\tok{odor}'' & ``\tok{Guardian}'' & ``\tok{Guardian}'' \\
24 & ``\tok{Guardian}'' & ``\tok{Guardian}'' & ``\tok{Guardian}'' \\
28 & ``\tok{Guardian}'' & ``\tok{Guardian}'' & ``\tok{Guardian}'' \\
\midrule
32 & ``\tok{Guardian}'' & ``\tok{Guardian}'' & ``\tok{Guardian}'' \\
\bottomrule
\end{tabular}

\caption{Top token of each reading after every fourth layer of OLMo3-7B at one position, where layer \NumLayersO{} is the output of the whole model.}
\label{tab:example}
\end{table}

\FloatBarrier
\section{Properties of Reference Frames}
\label{app:theory}

This appendix states in full the properties of reference frames. These frames express a hidden state in the native semantics of the model, through its own layers and with the context of the state. The appendix covers the two sides of a frame, how close a lens and an order can come to the model and the complexity of the orders.
Throughout, a state $x_\ell(t)$ is read with the states $x_\ell(s)$, $s < t$, of its prefix given, and $T^{(n)}_\ell$ denotes the transport of order $n$.

\subsection{Lexical Readout and Lexical Basis}

The lexical readout $b_\ell(v) = J_\ell^\top (w_v - \bar w)$ and the basis vector $s_\ell(v) = J_\ell^{+} s_v$ are the two sides of a frame.
The first is a covector, which measures how a small move of the state changes the centred logit of $v$.
The second is a vector. Among all moves $\delta$ of the state it is the smallest one whose first-order effect $J_\ell \delta$ on the normalised last state is closest to $s_v$, up to the ridge.
The same construction with $T^{(1)}_\ell$ in place of $T_\ell$ gives the lexical basis of the frame of order 1, \lrf{}.

The two sides are dual in the sense of frames rather than of bases.
The vocabulary is larger than the state, so the unembedding rows $w_v$ span the space of states without being a basis of it. The vectors $s_v = (W^\top W)^{-1} w_v$ are their canonical dual: $\sum_v s_v w_v^\top = I$, so a normalised last state is the sum of its logits times the vectors $s_v$.
To first order the same holds at layer $\ell$, up to the ridge in $J_\ell^{+}$ and the centring of the logits: $\sum_v s_\ell(v)\, (J_\ell^\top w_v)^\top = J_\ell^{+} J_\ell$, the projection onto the directions that reach the output.

At an intermediate layer the coefficient of a label cannot be taken as an inner product with the state itself.
By the chain rule, the pairing of a lexical readout with its state is
\begin{equation}
\bigl\langle b_\ell(v),\, x_\ell(t) \bigr\rangle = \frac{d}{d\alpha}\, c_v\bigl(\alpha\, x_\ell(t)\bigr)\Big|_{\alpha = 1} ,
\end{equation}
the response of the centred logit to a rescaling of the state, which the normalisations of the remaining layers largely remove.
It would equal $c_v$ only if the centred logit were homogeneous of degree one in the state, which the normalisations rule out, and its measured correlation with the centred logit is \CmpPairModelS{} and \CmpPairModelO{}.
The coefficient is therefore a value of the transported state and not a linear function of the state $x_\ell(t)$ at the layer.

\subsection{Coherence Bound and Exactness}

Let $\hat b_1, \dots, \hat b_m$ be the unit lexical readouts of one label at $m$ states of a layer, and let $r$ be a unit vector that does not depend on the state.
Its mean cosine with them is $\frac{1}{m}\sum_i r^\top \hat b_i = r^\top \bar b \leq \lVert \bar b \rVert$, with $\bar b$ the mean of the $\hat b_i$, and
\begin{equation}
\lVert \bar b \rVert^2 = \frac{1}{m^2}\sum_{i,j} \hat b_i^\top \hat b_j = \frac{1}{m} + \Bigl(1 - \frac{1}{m}\Bigr)\hat\kappa ,
\end{equation}
where $\hat\kappa$ is the mean cosine over pairs of distinct states. With states from many sequences, $\hat\kappa$ tends to the coherence $\kappa_\ell(v)$, which is measured over pairs from different sequences.
For many states the bound is $\sqrt{\kappa_\ell(v)}$, and it is attained by the mean direction $r = \bar b / \lVert \bar b \rVert$.
No lens, however it is fitted, can therefore exceed $\sqrt{\kappa_\ell(v)}$ on average, as the bound is a property of \lrfn{$\infty$} and not of an estimator. Since the square root is concave, the mean of these bounds over labels is at most $\sqrt{\kappa_\ell}$. A lens whose reading applies the final normalisation $N$ varies in direction with the state only through $N$, and in practice it also stays below this bound.
The bound concerns directions, not the divergence of a lens's reading from the model's output, which a lens fitted to that output can lower.

Write $h^{n}_k(t)$ for the state of position $t$ at layer $k > \ell$ in \lrfn{$n$}.
Order 0 replaces the attention of every layer by the typical attention output, and order $n \geq 1$ computes the layer's own attention over the prefix states of order $n-1$,
\begin{equation}
\begin{aligned}
h^{n}_k(t) &= F_k\bigl(h^{n}_{k-1}(t),\, a^{n}_k(t)\bigr) , \\
a^{n}_k(t) &= \mathrm{Att}_k\bigl(h^{n}_{k-1}(t);\, \{h^{n-1}_{k-1}(s)\}_{s<t}\bigr) ,
\end{aligned}
\end{equation}
with $h^{n}_\ell = x_\ell$ for every $n$.
By induction on $n$, the state of order $n$ equals the true state for its first $n$ layers, that is, $h^{n}_k(t) = x_k(t)$ for $k \leq \ell + n$.
This holds trivially for $n = 0$. If it holds for $n-1$, the attention of layer $k \leq \ell + n$ attends to prefix states of order $n-1$ at layer $k-1 \leq \ell + n - 1$, which are the true ones, and by induction on $k$ its query is the true state $x_{k-1}(t)$.
Two identities follow.
\lrfn{$n$} is exact wherever at most $n$ layers remain, $L - \ell \leq n$, so its reading, lexical readouts and lexical basis are those of \lrfn{$\infty$}. Further, for $n \geq L - \ell$ every state is the true one, so \lrfn{$\infty$} returns the model's output at every layer.
The orders differ from the model only through the attention between positions more than $n$ layers above the state.

\subsection{Complexity Analysis}

We count costs in forward passes, a measure independent of the hardware.
Calibration is one forward pass over the calibration text and one ridge regression per layer.
Reading the states of layer $\ell$ at order $n$ runs $n+1$ streams through the $L - \ell$ remaining layers. Reading every layer of a sequence, at all positions at once, therefore costs about $(n+1)(L+1)/2$ forward passes.
A fitted lens reads a state with one matrix product, which is cheaper per state, but the fit is not. The Jacobian lens needs the Jacobian of the last state with respect to the states of every layer. This takes one backward pass per dimension of the state for each of its $P$ prompts, about $P d$ backward passes in total.
With \lrf{}, a lens therefore becomes cheaper only after about $P d / (L+1)$ sequences have been read at every layer. The fit must be repeated for every checkpoint, while a frame repeats only its calibration.
Fitting the Jacobian lens took \CostRatioS{} and \CostRatioO{} times as long as calibrating the frames and reading the test sequences at orders 0 to 2.

\section{Discussion}
\label{app:discussion}

The framework rests on one idea: read a hidden state as the model reads it, through the layers that remain and with the context it sees.
Lenses read each layer with one fixed map and point clouds read no state at all, whereas a frame lets the layers that decide what a state means read it in the native semantics of the model itself.

A reference frame separates two questions that a lens answers at once: along which directions the model measures a label at a state, and how much of the label the state holds.
Existing methods appear as special cases. A lens is a frame that does not adapt to the context and is therefore bounded by the coherence of the native frame. A probe is a fixed lexical readout learned under supervision, and a point cloud is a set of states without a frame.
The native frame answers both questions exactly, and the orders make it informative by choosing what is withheld, so the contribution of each round of attention between positions can be measured. Both parts of the model's reading matter, the transport through the remaining layers and the attention to the prefix of the state.

The value of this view lies in what it makes comparable.
The vocabulary is shared by every layer, checkpoint and model of a family, and a frame expresses all of their states in it without a trained component whose faithfulness would need a separate argument. A frame calibrated anew in a single pass can therefore follow a model through training in the same units.
Early exit, hallucination detection and knowledge editing read hidden states through lenses or probes, and model compression matches the hidden states of a teacher. Each can instead read these states in the frame that the model itself uses to read them.

\end{document}